\documentclass[letterpaper]{article} 
\usepackage{aaai25}  
\usepackage{times}  
\usepackage{helvet}  
\usepackage{courier}  
\usepackage[hyphens]{url}  
\usepackage{graphicx} 
\usepackage{natbib}  
\usepackage{caption} 
\usepackage{algorithm}
\usepackage{algpseudocode}
\usepackage{subcaption} 
\usepackage{amsmath}
\usepackage{newfloat}
\usepackage{listings}
\usepackage{booktabs}
\usepackage{multirow}
\DeclareCaptionStyle{ruled}{labelfont=normalfont,labelsep=colon,strut=off} 
\floatstyle{ruled}
\newfloat{listing}{tb}{lst}{}
\floatname{listing}{Listing}
\title{VersusDebias: Universal Zero-Shot Debiasing for Text-to-Image Models\\via SLM-Based Prompt Engineering and Generative Adversary}
\nocopyright
\author {
    Hanjun Luo\equalcontrib,
    Ziye Deng\equalcontrib,
    Haoyu Huang\equalcontrib, 
    Xuecheng Liu,
    Ruizhe Chen,
    Zuozhu Liu\thanks{Corresponding author},
}
\affiliations{
    Zhejiang University\\
    hanjun.21@intl.zju.edu.cn, ziye.21@intl.zju.edu.cn, haoyuh.21@intl.zju.edu.cn, xuecheng.21@intl.zju.edu.cn, ruizhec.21@intl.zju.edu, zuozhuliu@intl.zju.edu.cn
}
\begin{document}

\maketitle

\begin{abstract}

With the rapid development of Text-to-Image (T2I) models, biases in human image generation against demographic social groups become a significant concern, impacting fairness and ethical standards in AI. Some researchers propose their methods to tackle with the issue. However, existing methods are designed for specific models with fixed prompts, limiting their adaptability to the fast-evolving models and diverse practical scenarios. Moreover, they neglect the impact of hallucinations, leading to discrepancies between expected and actual results. To address these issues, we introduce VersusDebias, a novel and universal debiasing framework for biases in arbitrary T2I models, consisting of an array generation (AG) module and an image generation (IG) module. The self-adaptive AG module generates specialized attribute arrays to post-process hallucinations and debias multiple attributes simultaneously. The IG module employs a small language model to modify prompts according to the arrays and drives the T2I model to generate debiased images, enabling zero-shot debiasing. Extensive experiments demonstrate VersusDebias's capability to debias any models across gender, race, and age simultaneously. In both zero-shot and few-shot scenarios, VersusDebias outperforms existing methods, showcasing its exceptional utility. Our work is accessible at https://github.com/VersusDebias/VersusDebias to ensure reproducibility and facilitate further research.

\end{abstract}

\section{Introduction}
Text-to-Image (T2I) models, as a crucial part of Artificial Intelligence Generated Content (AIGC) technology, are evolving at an extraordinary speed driven by open-source initiatives by companies such as Stability AI \cite{rombach2022high,luo2023latent,betker2023improving}, enabling even non-experts to create high-quality and photorealistic images from simple textual prompts. Nevertheless, similar to other generative models \cite{mehrabi2021survey, zhou2024bias}, the widespread adoption of T2I models has raised significant concerns regarding fairness and biases, particularly in images depicting human \cite{bansal2022well,luccioni2023stable}. Existing research highlights systematic biases within T2I models to specific social groups. For instance, T2I models primarily depict respectable occupations like lawyers and CEOs as white middle-aged men \cite{cho2023dall}, ignoring women and colored races.
\\
Some researchers have proposed their solutions to address the issue \cite{gandikota2024unified,schramowski2023safe,shrestha2024fairrag}. However, none of these methods fully resolve the two most crucial challenges that hinder the practical application of debiasing techniques. First, existing approaches are typically tailored to specific T2I models, commonly SD 1.5 \cite{rombach2022high}, which limits their adaptability, especially given the rapid evolution of T2I models driven by the adoption of distillation \cite{meng2023distillation} and the participation of communities. For instance, the primary base model iterates three times from SD1.5 to SDXL-L \cite{lin2024sdxl} within a year, while researchers introducing models with entirely new architectures \cite{pernias2023wurstchen,chen2023pixart}. Recent research still develops their methods based on outdated models, limiting their application prospects. Second, most methods lack zero-shot debiasing capabilities. For example, FairDiffusion \cite{friedrich2023fair} uses a predefined look-up table to recognize prompts and add gender description to improve fairness, failing to deal with prompts that fall outside the predefined table. Like other AIGC techniques, effective debiasing techniques need robustness and zero-shot debiasing abilities to handle diverse inputs, ensuring broader applicability.
\begin{figure}[H]
    \centering
    \includegraphics[width=0.4\textwidth]{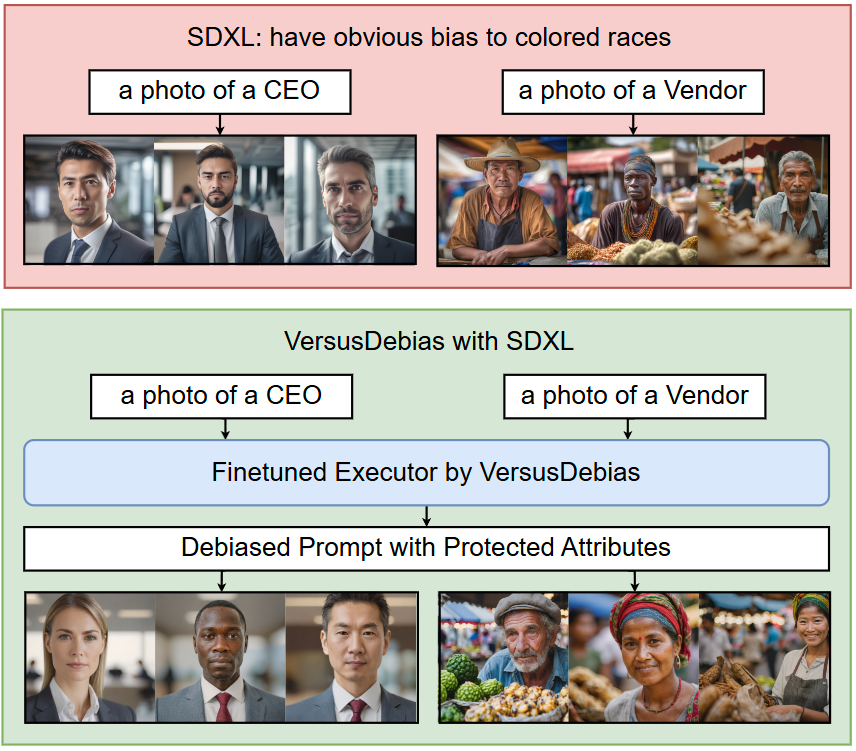}
    \caption{A simple generation workflow of VersusDebias.}
    \label{compare}
\end{figure}
\noindent
To address this, we propose VersusDebias, a universal and self-adaptive zero-shot debiasing framework for T2I models, consisting of an array generation (AG) module and an image generation (IG) module. A simple workflow is shown in Figure \ref{compare}. Referring the structure of generative adversarial networks \cite{goodfellow2020generative}, our AG module employs a generator and a discriminator to perform generative adversary. The discriminator employs a multi-modal large language model (MLLM) to achieve exceptional alignment performance. The AG module produces debiased attribute arrays corresponding to 2,060 prompts, which consist of protected attributes from the ignored social groups, providing reference for modifying the prompts. Compared to existing methods that add protected attributes according to fixed demographic proportions, debiased attribute arrays mitigate the hallucinations of T2I models, i.e., instances where models generate images that are inconsistent with the prompt, through post-processing. The IG module employs a fine-tuned small language model (SLM), i.e., LLM with fewer parameters, which can accurately complete named entity recognition (NER) in complex prompts and contextually add protected attributes according to debiased attribute arrays to generate debiased prompts, providing zero-shot debiasing capability. Our SLM achieves LLM-level performance on NER tasks, surpassing traditional NER methods, while significantly reducing computational resource consumption compared to LLMs, thereby striking a outstanding balance. VersusDebias is able to debias any models without retraining, avoiding computational overhead. Moreover, as a universal plug-in module, VersusDebias is equipped with ComfyUI API \cite{comfyui} and Websocket, enabling it to adapt to thousands of models compatible with ComfyUI by simply replacing the workflow. This structure, with the adaptive capabilities provided by generative adversary, enables VersusDebias to debias any T2I models. 
\\
To evaluate the effectiveness of our work, we compare VersusDebias with other methods from two perspectives. First, we compare the demographic bias of three general models with and without VersusDebias. Second, we compare VersusDebias with existing debiasing methods. All comparisons consider both zero-shot and few-shot scenarios with multiple evaluation datasets. The results demonstrate that VersusDebias effectively reduces biases in T2I models and significantly outperforms other methods.
\\
Our core contributions are summarized as follows:
\begin{itemize}
    \item We propose VersusDebias, a novel and universal framework to efficently decrease demographic biases in T2I models, which ensure fairness of the generated images.
    \item VersusDebias integrates numerous advanced technologies, including MLLM, SLM, prompt engineering, and generative adversary.
    \item VersusDebias implements zero-shot debiasing via the ComfyUI API and is transferable without retraining, showcasing its universality and promoting practical application of debiasing.
    \item Experiments show that VersusDebias outperforms existing methods in both zero-shot and few-shot debiasing situations, proving its effectiveness and robustness.
\end{itemize}

\section{Related Works}

\paragraph{Biases in T2I Models}
Over the last few years, T2I models like Stable Diffusion and DALL-E have seen increasing adoption \cite{ramesh2022hierarchical,esser2024scaling}. Many T2I models are able to generate high-quality images efficiently. Nonetheless, several studies prove that these models inherit biases from their training datasets \cite{chinchure2023tibet,wan2024survey}. Recent research classifies these biases into implicit generative bias and explicit generative bias \cite{luo2024faintbench}, corresponding to the sociological concepts of implicit bias \cite{pritlove2019good} and explicit bias \cite{fridell2013not}. Implicit generative bias refers to the phenomenon where, without specific instructions on protected attributes, T2I models tend to generate images that do not consist of demographic facts. For instance, when asked to generate "a photo of a surgeon", models tend to produce images featuring male surgeons. Explicit generative bias is a specific type of hallucination, referring to the phenomenon where T2I models tend to generate images that do not consist with prompts containing specific protected attributes. For example, when asked to generate "a photo of a rich black person" models may sometimes fail to correctly generate an image featuring a black person. In our work, we primarily address the issue of implicit generative bias and avoid the influence of explicit generative bias which disturbs existing methods.

\paragraph{Debiasing Based on Prompt Engineering}
Recent works attempts to debias T2I models at different levels, such as diffusion-level \cite{shrestha2024fairrag} and prompt-level \cite{clemmer2024precisedebias}. Diffusion-level methods interfere with the image generation process, which can lead to uncontrollable side effects on the quality and style of the final image. Moreover, these methods can only be applied to diffusion-based models and are not suitable for transformer-based models like DALL-E3 \cite{betker2023improving}. In contrast, debiasing based on prompt engineering, as a post-processing method, is more adjustable and more broadly applicable with less impact on image quality. These methods typically performs NER to identify the bias-inducing parts of the prompt and modifies them according to predefined proportions. However, these methods also face serious challenges. First, traditional prompt engineering methods rely on the assumption that the model correctly generate images based on the prompt, which highlights the problem of hallucinations, especially those caused by explicit generative bias. Second, prompt engineering might have unintended side effects on unrelated attributes. For example, using SDXL-Turbo \cite{sauer2023adversarial}, the proportion of women generated with the prompt "a photo of a South Asian tennis player" is significantly lower than that generated with the prompt "a photo of a tennis player" \cite{luo2024bigbench}. Lastly, the severity of these issues varies across different models, meaning that traditional methods, even if they address the first two issues, can only be applied to specific models, greatly limiting their application. In contrast, our method employs an innovative AG module that addresses these problems, providing unprecedented practicality.

\begin{figure*}[t]
    \centering
    \includegraphics[width=0.85\textwidth]{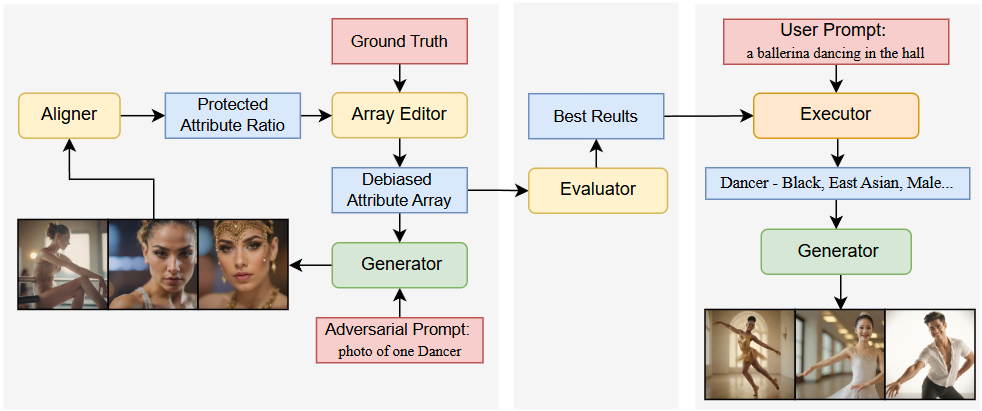}
    \caption{Pipeline of VersusDebias. The yellow boxes represent the components of the Discriminator. The three gray rectangles distinguish the three main stages. The blue rectangles represent data that change with the T2I model, while the red rectangles represent data that is independent of the T2I model.}
    \label{artitecture_sum}
\end{figure*}

\paragraph{Development of SLM and MLLM}
Due to the tremendous success of ChatGPT-4 \cite{achiam2023gpt}, LLMs, especially MLLMs, garner significant attention and rapidly evolve. The improved accuracy of MLLMs addresses the significant shortcomings of traditional alignment models like CLIP \cite{radford2021learning} and BLIP-2 \cite{li2023blip} in recognizing race and age \cite{chinchure2023tibet}, providing essential support for the implementation of the alignment module in our discriminator. On the other hand, the development of edge AI attracts more researchers to focus on SLMs \cite{bai2023qwen,zhang2024tinyllama}, which retain most of the original performance of LLMs while significantly reducing computational resource consumption. Compared to traditional NER methods, LLM-based methods exhibit higher accuracy and zero-shot capabilities \cite{clemmer2024precisedebias}, which also indicates the potential of SLMs in NER. In VersusDebias, we utilize a recent SLM and a MLLM to achieve outstanding debiasing performance and accuracy with minimal computational resources.

\section{Method}
\subsection{Overview}
In this section, we introduce the pipeline of VersusDebias. A brief overview is shown in Figure \ref{artitecture_sum}. VersusDebias consists of a AG module, a IG module, and their datasets. 
\\
To debias a model, VersusDebias first uses prompts without protected attributes from the dataset to drive the generator to generate images, which are then conveyed into the discriminator. The discriminator employs a MLLM to perform semantic alignment on these images, extracting information on protected attributes. Due to the lack of statistic, we only consider the debiasing of gender, race, and age of the depicted persons. The alignment results are averaged to determine the generative demographic proportions of protected attributes for each prompt. The discriminator compares the proportions of each gender, race, and age group to the ground truth and modifies the prompts by adding the ignored protected attributes, which ensures that the generator produces more images featuring these ignored social groups in the next iteration. The added protected attributes are recorded in the protected attribute array corresponding to each prompt. The newly images generated from modified prompts are then conveyed into the discriminator again, and the process is repeated until the specified number of epochs.
\\
The discriminator then evaluates the generation results for each prompt at each epoch, calculating the cosine similarity between the generative proportions and the ground truth to quantify the conformity between them. The protected attribute array with the highest cosine similarity for each prompt across different epochs is then combined into a single dictionary, referred to as the best results, and conveyed to the IG module.
\\
For generating debiased images with the IG module, the executor uses the fine-tuned SLM to do NER accurately by extracting the potentially bias-inducing part from the input user prompt. It then selects an appropriate protected attribute based on the best results and adds it to the original prompt. Finally, the generator uses the processed prompt to generate debiased images.

\subsection{Generator}
As a universal debiasing framework based on prompt engineering, VersusDebias can use any model in its generator, including diffusion-based models \cite{nichol2021glide,sauer2023adversarial,esser2024scaling,lin2024sdxl,song2024sdxs} and transformer-based models \cite{dayma2021dalle,ding2022cogview2,ramesh2022hierarchical,chen2023pixart}. To lower the barrier of VersusDebias and attract more researchers to debiasing studies, we develop an interface with ComfyUI, which is a widely-used GUI for thousands of T2I models based on Stable Diffusion. ComfyUI only requires changes to the input workflow file to adapt to various models and is equipped with a comprehensive API for external program calls. Additionally, due to its openness and flexibility, some models that use new architectures, such as Playground \cite{li2024playground}, PixArt \cite{chen2024pixartd}, and Stable Cascade, are also compatible with ComfyUI by simply installing specific extension nodes. In terms of communication, VersusDebias uses Websocket to monitor the status of ComfyUI and convey instructions. This approach not only achieves a fully automated debiasing process but also allows both systems to operate in independent environments, ensuring optimal compatibility.
\\
In IG module, the generator uses the prompt processed by the executor to generate images based on user input. In AG module, the generator uses prompts from our dataset. The creation process for the dataset is as follows. Based on the 607 detailed occupations classified by the standard occupational classification of the U.S. \cite{BLS2024}, we consolidate occupations that are typically not distinguished in everyday contexts, resulting in 103 occupations. We also add descriptions of image quality and character states with ChatGPT-4o, generating 20 prompts for each occupation and totaling 2,060 prompts. The detailed process is provided in the supplementary material.

\subsection{Discriminator}
As the core part of the AG module, the discriminator consists of three components: the aligner, the array editor, and the evaluator. In the following sections, we provide a detailed explanation of the implementation of them.

\paragraph{Aligner}
The aligner's function is to align protected attributes with individuals in the images and compute the average of the responses of each occupation. We use the fine-tuned InternVL-4B-1.5 \cite{chen2024far} from BIGbench for alignment, which is trained with the dataset from FairFace \cite{karkkainen2021fairface} and reportedly achieves an accuracy of 97.93\%. Using race alignment as an instance, the aligner asks the model "Tell me the race of the main person in the image, White, Black, East Asian, South Asian, or unknown", and store the response except "unknown". The aligner clean the history and tries again if "unknown", and skip the image if still "unknown" persists, assuming the T2I model failed to generate human image. We do a test with results shown in the supplementary material, indicating that this model meeting the requirements of VersusDebias. 

\paragraph{Array Editor}
\label{size_why}
The function of the array editor is comparing the generative demographic proportions to the ground truth and adding the underrepresented protected attributes to the prompt. To achieve this, the array editor generates a new dictionary by subtracting the ground truth values from the actual generative proportions. The keys in this dictionary are the occupations, with negative values indicating that the generated data is below the actual data and positive values indicating the opposite. For all protected attributes with negative differences, the array editor adds these underrepresented attributes to the attribute array. The length of the attribute array $L$ can be adjusted based on the requirements for speed and accuracy, but it should generally be maintained at 4 times the number of epochs. This is because there are a total of nine protected attributes, and assuming an equal distribution of positive and negative differences, approximately 4.5 attributes are added each time. Allowing for a buffer, the multiplier is set to 4 as a floor value, and $L$ is at least 5 as a ceiling value. The array editor performs this operation once per epoch until the limit is reached. A detailed algorithm is shown in Algorithm \ref{editor}.

\begin{algorithm}[H]
\caption{Array Editing Algorithm}
\label{editor}
\textbf{Input}: Ground truth values $G$, actual generative proportions $A$, number of epochs $E$, length multiplier $m=4$ \\
\textbf{Output}: Adjusted attribute array
\begin{algorithmic}[1] 
\State Initialize dictionary $D \gets \{\}$
\For{each prompt $p$ in $G$}
    \State $D[p] \gets A[p] - G[p]$
\EndFor
\State Initialize attribute array $attrArray \gets []$
\State Set limit $limit \gets 9 \times E \times m$
\For{epoch $e$ from 1 to $E$}
    \For{each attribute $a$ in $D$}
        \If{$D[a] < 0$} \State Append $a$ to $attrArray$ \EndIf
    \EndFor
    \If{length($attrArray$) $\geq$ $limit$} \State \textbf{break} \EndIf
\EndFor
\State \textbf{return} $attrArray$
\end{algorithmic}
\end{algorithm}
\noindent
The ground truth uses the same occupational classification as the prompt set. For race, we employ the four-category classification from FairFace \cite{karkkainen2021fairface}, distinguishing race into White, Black, East Asian, and South Asian. For age, we classify it into three stages: young (0-30), middle-aged (31-60), and elderly (60+), based on common age classifications. For gender, due to the cognitive limitations and the diverse of sexual minority identities, we classify gender into male and female, which limits our method to debias for sexual minorities. For race proportions, we use the global racial population data from the UN \cite{un_population_2022}, as VersusDebias is designed for worldwide users. For gender and age, the proportions for practitioners for a occupation are calculated by dividing the number of individuals of a specific group by the total number of individuals within all detailed occupation belonging to the occupation, based on the data from the BLS \cite{BLS2024a,BLS2024b}, due to a lack of global data on occupation practitioners and this data also following the SOC.
\paragraph{Evaluator}
The evaluator provides a comprehensive metric for evaluating the overlap between the generated results and the ground truth when selecting the best-performing debiased attribute arrays. Since gender, race, and age each have 2-4 protected attributes, we treat them as vectors and use cosine similarity to determine the similarity between the generated results and the ground truth:
\begin{equation}
\label{cos}
S_{k} = \frac{\sum_{i=1}^{n_k} p_{ki} \cdot q_{ki}}{\sqrt{\sum_{i=1}^{n_k} p_{ki}^{2}} \cdot \sqrt{\sum_{i=1}^{n_k} q_{ki}^{2}}},
\end{equation}
where $S_{k}$ is the cosine similarity for the protected attribute $k$ of a occupation, $p_ki$ and $q_ki$ are the generative proportion and real demographic proportion of the $i$ attribute, and $n_k$ is the total number of the attributes for this protected attribute.\\
After calculating the cosine similarity for these three attributes separately, the evaluator computes a weighted average to obtain the cumulative cosine similarity. We set the weights for gender and race at 0.35 and the weight for age at 0.3, which reflects the general consensus that gender and racial discrimination is more common and severe. The weights can be adjusted based on different requirements:
\begin{equation}
S_{c} = \sum_{k \in \{\text{gender}, \text{race}, \text{age}\}} w_k \cdot S_k,
\end{equation}
where $S_{c}$ is the cumulative cosine similarity for a occupation, $w_k$ is the weight for the protected attribute $k$.\\
Finally, the evaluator selects the best-performing debiased attribute arrays based on the cumulative cosine similarity and combines them into a single dataset called "best results".
\subsection{Executor}
\begin{figure}[H]
    \centering
    \includegraphics[width=0.35\textwidth]{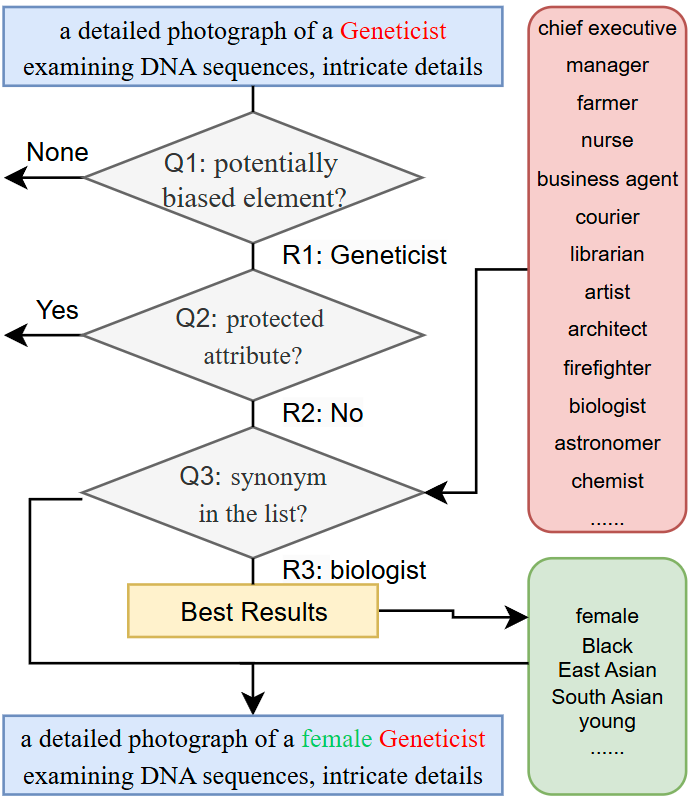}
    \caption{The pipeline of the executor. The blue rectangles represents the prompt, the red box represents the list of all occupations included in the prompt set, and the green box represents the debiased attribute arrays for R3.}
    \label{exe}
\end{figure}
\noindent
The pipeline of the executor is briefly shown in Figure \ref{exe}. The main component of the executor is a fine-tuned SLM. We use the SLM to accurately perform the following tasks in sequence: first, do NER to analyze practical prompts and recognize the potentially biased part; second, determine whether the prompt already includes protected attributes; third, identify a occupation in the occupation list that has the closest meaning to response 1. The executor locates the debiased attribute arrays corresponding to response 3 from the best results, randomly selects a protected attribute from the generated debiased attribute array, and adds the selected protected attribute to the prompt.
\\
Due to the relatively fixed and simple nature of the tasks performed, we use a SLM instead of traditional LLMs to accomplish the aforementioned three tasks. SLMs are characterized by their faster operational speed and less resource consumption. However, they are subject to limitations in scale, which can lead to hallucinatory issues and less precise knowledge memory. To ensure accuracy, we develop a specialized dataset to fine-tune the model, thereby enhancing its capability to execute these tasks, particularly in the retrieval of synonyms. The dataset $D_{SLM}$ we used to train the SLM consists of 500 dialogue sets, with each set containing three rounds of queries $q$ and expected responses $r$, which is formulated as follows:
\begin{equation}
\label{dataset}
D_{SLM} = \{ (q_{i,k},r_{i,k}) \mid i \in [1,500], k \in [1,3] \}
\end{equation}
In the first round, $q_1$ asks the model to identify the potentially biased part in a given prompt. $r_1$ is "none" or the identified element, thus training the model's ability of extracting. To closely simulate practical application scenarios, we use Promptomania \cite{promptomania} to generate one prompt for each of the 103 occupations in the prompt set. Additionally, we use the official US occupational classification method \cite{SOC2018} and thesaurus \cite{thesaurus} to collect synonyms and variations for the occupations, resulting in a total of 380 queries. We test these prompts with SDXL \cite{podell2023sdxl} and manually revise prompts that produced poor image results to improve their quality. To ensure the SLM does not misidentify potentially biased parts, we also include 120 queries containing prompts that solely describe landscapes and objects to train the model's judgment capabilities. In the second round, $q_2$ asks the model to determine whether the prompt originally contains a protected attribute, with $r_2$ being either "yes" or "no". We included 260 sets without protected attributes and 120 sets with protected attributes to comprehensively train the model. In the third round, $q_3$ asks the model to select a synonym from the occupation list for $r_1$ if it's not "none" while $r_3$ is the chosen synonym.

\section{Experiment}
In this section, we initially present the comparative experiments conducted in the selection of SLMs. Subsequently, we summarize the performance of VersusDebias in debiasing various T2I models for both few-shot and zero-shot scenarios and investigate the influence of the attribute array length $L$. Finally, we compare VersusDebias with several baselines. 
\subsection{SLM Comparison}
We select three state-of-the-art SLMs, Qwen2-1.5B \cite{yang2024qwen2}, MiniCPM-2B \cite{hu2024minicpm}, and Gemma-2B \cite{gemma2024}. For comparative analysis, we choose representative LLMs deployable on a single GPU, Llama2-7B \cite{touvron2023llama} and Qwen2-7B, and the representative commercial large model, ChatGPT-4o \cite{achiam2023gpt}. All models except ChatGPT-4o are fine-tuned using our dataset $D_{SLM}$.
\\
We conduct the analysis by developing a test set comprising 100 dialogue sets, consistent with the format of the training set, which consists of 52 dialogues describing humans without protected attributes, 24 describing humans with protected attributes, and 24 not involving human descriptions. With the test set, we evaluate the five models and calculated the accuracy by comparing the responses to the results of one human annotator undergoing specific training to evaluate the dialogues and occupations involved. We provide the datasheet in our respiratory for verification. Particularly, if $r_1$ can be categorized under multiple occupations in the element list, all of the occupations are correct for $r_3$. For instance, if $r_1$ is "robotics engineer," then $r_3$ is correct if its value is "electrical engineer," "mechanical engineer," or "computer programmer". The performance of models is detailed in Table \ref{exe_compare}. Qwen2-1.5B, while being the fastest in terms of speed with the usage of 5.4GB VRAM, achieves an accuracy that is slightly better than Llama2-7B and significantly higher than the other two SLMs, making it our choice. Furthermore, the performance of fine-tuned Qwen2-1.5B and Llama2-7B approaches ChatGPT-4o, which attests to the validity of our dataset design. 
\begin{table}[H]
    \centering
    \begin{tabular}{lccccc}
        \toprule
        \textbf{Model} & $r_1$ & $r_1$-w/o & $r_2$ & $r_3$ & Speed \\
        \midrule 
        ChatGPT-4o & 100 & 100 & 100 & 98.65 & - \\
        Qwen2-7B & 100 & 100 & 98.65 & 98.65 & 2.05 \\
        Llama2-7B & 94.59 & 96.15 & 98.65 & 97.3 & 1.93 \\
        MiniCPM-2B & 90.54 & 88.46 & 95.95 & 91.89 & 2.67 \\
        Gemma-2B & 82.43 & 92.31 & 95.95 & 83.78 & 2.24 \\
        \midrule
        \textbf{Qwen2-1.5B} & 97.3 & 100 & 98.65 & 95.95 & 4.23 \\
        \bottomrule
    \end{tabular}
    \caption{The $r_1$ column reflects the accuracy of the models when processing prompts that describe humans, while the $r_1$-w/o column reflects the accuracy for prompts without human descriptions. The Speed column indicates the average number of prompts processed per second with a RTX3090.}
    \label{exe_compare}
\end{table}
\subsection{Debiasing Performance}

\paragraph{Few-shot}
For few-shot scenarios, we utilize ChatGPT-4o to generate 5 prompts for each of the 103 professions in the dataset and each prompt is used to generate 8 images. The format of these prompts imitates the pratical prompts. We employ these prompts to generate images and calculated the cosine similarity against the ground truth in the dataset. We calculate the cosine similarity between the results and ground truth of three T2I models, both with and without VersusDebias at a $L$ of 20. The two diffusion-based models are Stable Diffusion 1.5 \cite{rombach2022high} and Stable Diffusion XL \cite{podell2023sdxl}, while the transformer-based model is Pixart-$\Sigma$ \cite{chen2024pixart}. These models are abbreviated as SDv1, SDXL, and PixArt in the following part. The result is shown in Table \ref{few-shot}. The results fully demonstrate VersusDebias's capability to effectively debias models of various principles in few-shot scenarios, especially in race and age where the models originally performed poorly. It is notable that PixArt, which initially performed worse, outperforms the originally better-performing SDv1 after debiasing. Upon reviewing the generated images, we find that this is because PixArt has a better capability to follow the guidance of the prompt. This highlights the importance of this capability for prompt-based debiasing.
\begin{table}[H]
    \centering
    \begin{tabular}{lcccc}
        \toprule
        \textbf{Model} & \textbf{Gender} & \textbf{Race} & \textbf{Age} & \textbf{Total} \\
        \midrule
        SDv1 & 93.71 & 70.77 & 60.19 & 75.63 \\
        SDv1-VD & 95.23 & 86.75 & 81.65 & 88.19 \\
        \midrule
        SDXL & 89.15 & 65.52 & 78.85 & 77.79 \\
        SDXL-VD & 96.24 & 85.29 & 91.44 & 90.97 \\
        \midrule
        PixArt & 86.42 & 60.25 & 72.65 & 73.13 \\
        PixArt-VD & 92.83 & 83.26 & 92.75 & 89.46 \\
        \bottomrule
    \end{tabular}
    \caption{Qualitative results in few-shot scenarios.}
    \label{few-shot}
\end{table}
\paragraph{Zero-shot}
For zero-shot scenarios, we design the evaluation based on BIGbench \cite{luo2024bigbench}. Firstly, we exclude the prompts from BIGbench if the occupations in the prompts overlap with our prompt set, and use the remaining 122 occupations and 2440 prompts to generate 8 images for each prompt. Subsequently, we calculate the cosine similarity. Like the evaluation of few-shot debiasing capabilities, we also conducted comparative experiments for SDv1, SDXL, and PixArt with and without VersusDebias. The result is shown in Table \ref{zero-shot}. For zero-shot scenarios, VersusDebias also achieves outstanding performance.
\begin{table}[H]
    \centering
    \begin{tabular}{lcccc}
        \toprule
        \textbf{Model} & \textbf{Gender} & \textbf{Race} & \textbf{Age} & \textbf{Total} \\
        \midrule
        SDv1 & 91.36 & 68.13 & 53.94 & 72.00 \\
        SDv1-VD & 95.12 & 87.59 & 77.21 & 87.11 \\
        \midrule
        SDXL & 88.06 & 67.53 & 74.18 & 76.71 \\
        SDXL-VD & 93.50 & 88.38 & 87.23 & 89.83 \\
        \midrule
        PixArt & 85.92 & 58.87 & 70.25 & 71.75 \\
        PixArt-VD & 90.73 & 86.16 & 87.21 & 88.07 \\
        \bottomrule
    \end{tabular}
    \caption{Qualitative results in zero-shot scenarios.}
    \label{zero-shot}
\end{table}

\begin{table*}[t]
\centering
\begin{tabular}{@{}l|ccc|c|ccc|c|ccc|c@{}}
\toprule
\multirow{2}{*}{\bf Model} & \multicolumn{3}{c|}{\bf BIGbench} & \multirow{2}{*}{\bf total} & \multicolumn{3}{c|}{\bf VersusDebias} & \multirow{2}{*}{\bf total} & \multicolumn{3}{c|}{\bf PreciseDebias} & \multirow{2}{*}{\bf total} \\
\cmidrule(lr){2-4} \cmidrule(lr){6-8} \cmidrule(lr){10-12}
& gender & race & age &  & gender & race & age &  & gender & race & age &  \\
\midrule
SDv1 & 91.36 & 68.13 & 53.94 & 72.00 & 93.71 & 70.77 & 60.19 & 75.63 & 85.34 & 58.90 & 51.78 & 66.02 \\  
FairDiffusion & 95.90 & 73.80 & 64.14 & 78.64 & 95.12 & 69.18 & 62.23 & 76.17 & 92.19 & 61.81 & 52.63 & 69.69 \\ 
PreciseDebias & 86.92 & 95.56 & 54.14 & 80.11 & 84.55 & 96.17 & 58.21 & 80.72 & 77.83 & 80.63 & 58.79 & 73.10 \\
\midrule
VersusDebias & 95.12 & 87.59 & 73.21 & 85.91 & 95.23 & 86.75 & 81.65 & 88.19 & 91.69 & 72.73 & 75.41 & 80.17 \\ 
\bottomrule
\end{tabular}
\caption{VersusDebias exhibits the best overall performance and balanced results across attributes, demonstrating its comprehensiveness. We provide more detailed experimental results and further discuss the differences in the supplementary material.}
\label{cumulative}
\end{table*}

\begin{figure}[H]
    \centering
    \includegraphics[width=0.45\textwidth]{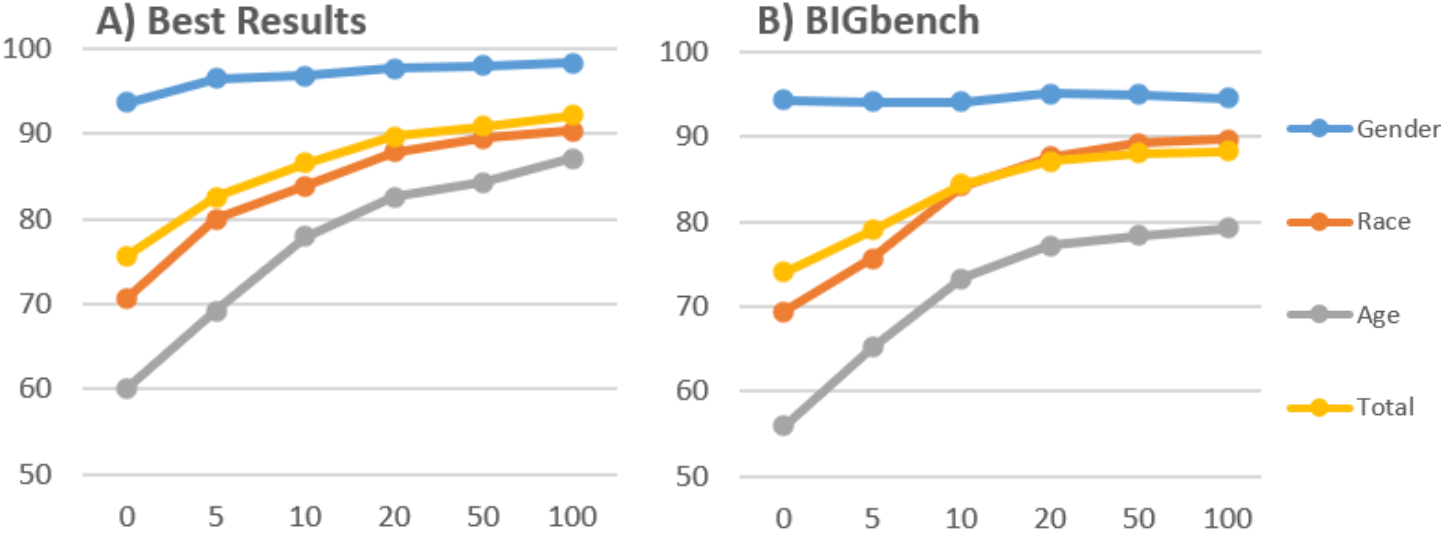}
    \caption{Performance curve with the increase of $L$.}
    \label{epoch}
\end{figure}

\paragraph{Influence of Array Length $L$}
Due to VersusDebias's reliance on randomly selecting protected attributes from debiased attribute arrays, the probability of selecting a protected attribute is quantized. Therefore, a larger attribute array can result in smoother variations in generative demographic proportions, improving accuracy. As explained in Section \ref{size_why}, the length of the debiased attribute array $L$ is greater than or equal to 5. We conduct comparative experiments with $L$ set to 5, 10, 20, 50, and 100, corresponding to the epochs being 2, 3, 5, 13, and 25, respectively. The T2I model used in the experiment is SDv1. The result is shown in Figure \ref{epoch}. 
\begin{figure}[H]
    \centering
    \includegraphics[width=0.45\textwidth]{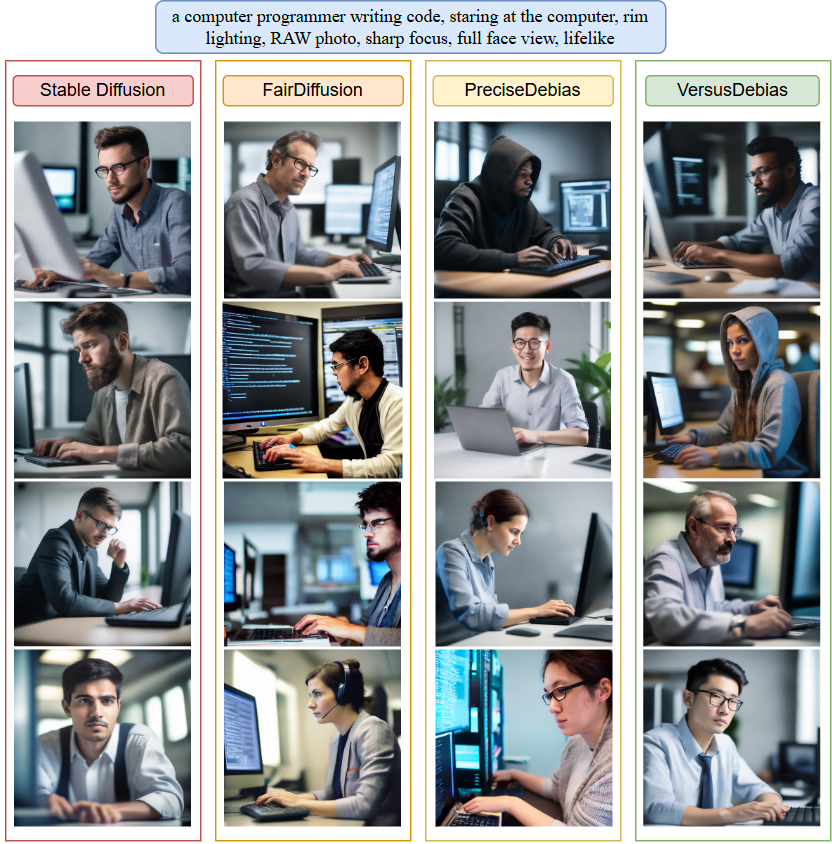}
    \caption{With VersusDebias, the generated results encompass all genders, races, and ages, and proportions closely matches the ground truth.}
    \label{exp}
\end{figure}
\noindent
The x-axis of the graph represents $L$, and the y-axis represents cosine similarity in the best results for few-shot or the results of BIGbench for zero-shot. An array size "0" indicates the performance without VersusDebias. The results show that larger array lengths effectively enhance the performance, with a clear marginal effect. Therefore, we set $L$ to 20 in other experiments to achieve a balance between performance and speed. Detailed computational resource consumption is provided in the supplementary material.

\subsection{Comparison with Baselines}

As no prior universal debiasing methods have been developed, we select two prompt-level existing methods based on SDv1, FairDiffusion \cite{friedrich2023fair} and PreciseDebias \cite{clemmer2024precisedebias}. We also use SDv1 for VersusDebias in this part of the experiment to control variables. FairDiffusion uses a look-up table to do NER and adds gender descriptions based on predefined probabilities for different occupations. PreciseDebias employs a fine-tuned Llama2-7B model to do NER and adds gender and racial descriptions according to fixed general probabilities. In the contrast, VersusDebias debiases gender, age, and race simultaneously via the AG module without knowing all specific proportions. To ensure the fairness, we replace the original racial proportions in PreciseDebias with the proportions from the ground truth and include additional test based on the dataset of PreciseDebias. Table \ref{cumulative} shows the results and Figure \ref{exp} provides a visualized example.

\section{Limitation and Future Work}
Although VersusDebias significantly outperforms existing debiasing methods, our framework still need to be improve in some aspects. First, although the fine-tuned InternVL-4B-1.5 demonstrates excellent alignment, particularly in racial alignment, it still cannot achieve perfect alignment. This limitation reduces the accuracy of the discriminator, which in turn affects the performance of the discriminator. A better model that is specifically designed for human classification is worth investigating. Secondly, as a post-processing method, VersusDebias fails to handle extreme cases of hallucinations, such as when the model completely fails to generate images that match the prompt "an Asian husband and a white wife" \cite{verge2024instagram}. This issue needs model-level improvement. Lastly, since VersusDebias only solves the problem of implicit bias, explicit bias remains to be addressed in future work. We hope our framework can be helpful for future research about explicit bias.

\section{Conclusion}
This paper introduces VersusDebias, a universal debiasing framework for T2I models, which enables precise and efficient debiasing for any T2I models by prompt engineering. To achieve this, we design a novel AG module to generate debiased attribute arrays for specific models and apply a advanced fine-tuned SLM to do NER and edit prompts according to the arrays. The edited prompts are used in T2I models to generate images with less demographic biases. Our AG module addresses the issues of hallucinations and side effects caused by prompt engineering, which are not handled by previous prompt-level methods, significantly improving the performance of our method. We also use ComfyUI API to lower the barriers of debiasing. VersusDebias surpasses existing methods in both zero-shot and few-shot scenarios, while maintaining the models' image style and quality. We hope that VersusDebias can promote the application of debiasing methods in T2I models, for a fairer AIGC community.

\bibliography{refer}
\newpage
\setcounter{secnumdepth}{0}
\section{Prompt Collection}
We follow the pipeline of BIGbench \cite{luo2024bigbench} to construct our dataset for the AG module. We add descriptions to the 103 occupations, generating 20 prompts for each occupation and totaling 2,060 prompts.
\\
Each of the 2,060 prompts is composed of three parts. The first part specifies an occupation. The second part has two components: one describing the environment around the person, and the other focusing on the person's expression, demeanor, or accessories. Since these prompts need to conform to numerous identity prompts, we use ChatGPT-4o \cite{achiam2023gpt} to generate them instead of random programs. The third part enhances the realism of the image, consisting of four components, each used to ensure the clarity of facial features, sharpen the overall clarity, maintain a realistic style, and ensure that the image depicts only one person. This part uses a random program to assign prompts from the predefined list to each of the four components.

\section{Evaluation of Aligner}
To evaluate the performance of the MLLM in our framework, we do a test at a sample of 50 images, indicating that this model exhibits high accuracy, meeting the requirements of VersusDebias. The dataset is provided in our repository for vindication.
\begin{table}[H]
    \centering
    \begin{tabular}{lccccc}
        \toprule
        \textbf{Model} & Gender & Race & Age \\
        \midrule
        \textbf{fine-tuned InternVL} & 50/50 & 49/50 & 48/50 \\
        \bottomrule
    \end{tabular}
    \caption{Summary of the accuracy of the fine-tuned Mini-InternVL-4B-1.5 \cite{chen2024internvl}. The number before the slash represents the correctly identified images, while the number after the slash represents the total number of tested images.}
    \label{internvl}
\end{table}
\section{Detailed Influence of Array Length $L$}
Due to VersusDebias's reliance on randomly selecting protected attributes from debiased attribute arrays, the probability of selecting a protected attribute is quantized. Therefore, a larger attribute array can result in smoother variations in generative demographic proportions, improving accuracy. However, it also requires more computational resource. In this section, we provide detailed experimental results. 
\subsection{Computational Resource Consumption}
\begin{table}[H]
    \setlength{\tabcolsep}{1mm}
    \centering
    \begin{tabular}{lccc}
        \toprule
        \textbf{$L$} & \textbf{Epoch} & \textbf{time (hours)} \\
        \midrule
        5 & 2 & 3.57 \\
        10 & 3 & 5.46 \\
        20 & 5 & 8.89 \\
        50 & 13 & 23.15 \\
        100 & 25 & 42.68 \\
        \bottomrule
    \end{tabular}
    \caption{Epochs and GPU hours for different array sizes. The GPU is one RTX3090.}
    \label{tab:time_table}
\end{table}
\subsection{Quantitative Results}
\begin{table}[H]
    \centering
    \begin{tabular}{lcccc}
        \toprule
        \textbf{$L$} & \textbf{Gender} & \textbf{Race} & \textbf{Age} & \textbf{Total} \\
        \midrule
        \textbf{0} & 93.71 & 70.77 & 60.19 & 75.63 \\
        \textbf{5} & 96.45 & 80.02 & 69.20 & 82.52 \\
        \textbf{10} & 96.78 & 83.85 & 77.89 & 86.59 \\
        \textbf{20} & 97.70 & 87.83 & 82.54 & 89.70 \\
        \textbf{50} & 97.98 & 89.49 & 84.27 & 90.90 \\
        \textbf{100} & 98.30 & 90.34 & 87.06 & 92.14 \\
        \bottomrule
    \end{tabular}
    \caption{Quantitative results at different $L$ for few-shot scenarios, i.e., the cumulative cosine similarity of best results of AG module.}
    \label{tab:new_final_scores_rounded}
\end{table}
\begin{table}[H]
    \centering
    \begin{tabular}{lcccc}
        \toprule
        \textbf{$L$} & \textbf{Gender} & \textbf{Race} & \textbf{Age} & \textbf{Total} \\
        \midrule
        \textbf{0} & 94.34 & 69.44 & 55.94 & 74.11 \\
        \textbf{5} & 94.06 & 75.73 & 65.20 & 78.99 \\
        \textbf{10} & 94.10 & 84.22 & 73.35 & 84.42 \\
        \textbf{20} & 95.12 & 87.59 & 77.21 & 87.11 \\
        \textbf{50} & 95.02 & 89.30 & 78.40 & 88.03 \\
        \textbf{100} & 94.54 & 89.74 & 79.28 & 88.28 \\
        \bottomrule
    \end{tabular}
    \caption{Quantitative results at different $L$ for zero-shot scenarios with the dataset of BIGbench.}
    \label{tab:final_scores_rounded}
\end{table}

\section{Detailed Comparison with Baselines}
\subsection{Limitations of Baselines}
As no prior universal methods have been developed, we select two existing methods based on the SDv1 as our baselines, FairDiffusion \cite{friedrich2023fair} and PreciseDebias \cite{clemmer2024precisedebias}. We also use SDv1 for VersusDebias in this part of the experiment to control variables. These methods are composed of two parts: one for performing the NER task and the other for adding a protected attribute before the identified occupation based on predefined probabilities.
\\
FairDiffusion uses a look-up table to do NER and adds gender descriptions based on predefined probabilities for different occupations. Its limitations include a reliance on a fixed look-up table for predefined occupations in NER tasks, leading to a low recognition rate. Moreover, it only addresses gender bias, which significantly limits its effectiveness.
\\
PreciseDebias employs a fine-tuned Llama2-7B model to do NER and adds gender and racial descriptions according to fixed general probabilities. While the LLM enhance NER performance and flexibility, the 7B model's 20GB VRAM requirement limits its widespread use. Additionally, fixed probabilities based on U.S. ethnic population proportions from the BLS \cite{BLS2024}, shown in Table \ref{precise} can be counterproductive, creating a larger gap from the actual statistical data. For example, the number of female nurses far exceeds that of male nurses, and strictly generating based on equal proportions can actually increase bias. Lastly, it does not consider age-related bias, which is a notable limitation.
\begin{table}[H]
    \centering
    \begin{tabular}{lcccc}
        \toprule
        \textbf{} & \textbf{white} & \textbf{black} & \textbf{asian} & \textbf{hispanic} \\
        \midrule
        \textbf{male} & 38.5 & 6.3 & 3.3 & 9.2 \\
        \textbf{female} & 38.5 & 6.3 & 3.3 & 9.2 \\
        \bottomrule
    \end{tabular}
    \caption{Demographic distribution in PreciseDebias.}
    \label{precise}
\end{table}

\subsection{Visualized Results}
\begin{figure}[H]
    \centering
    \includegraphics[width=0.47\textwidth]{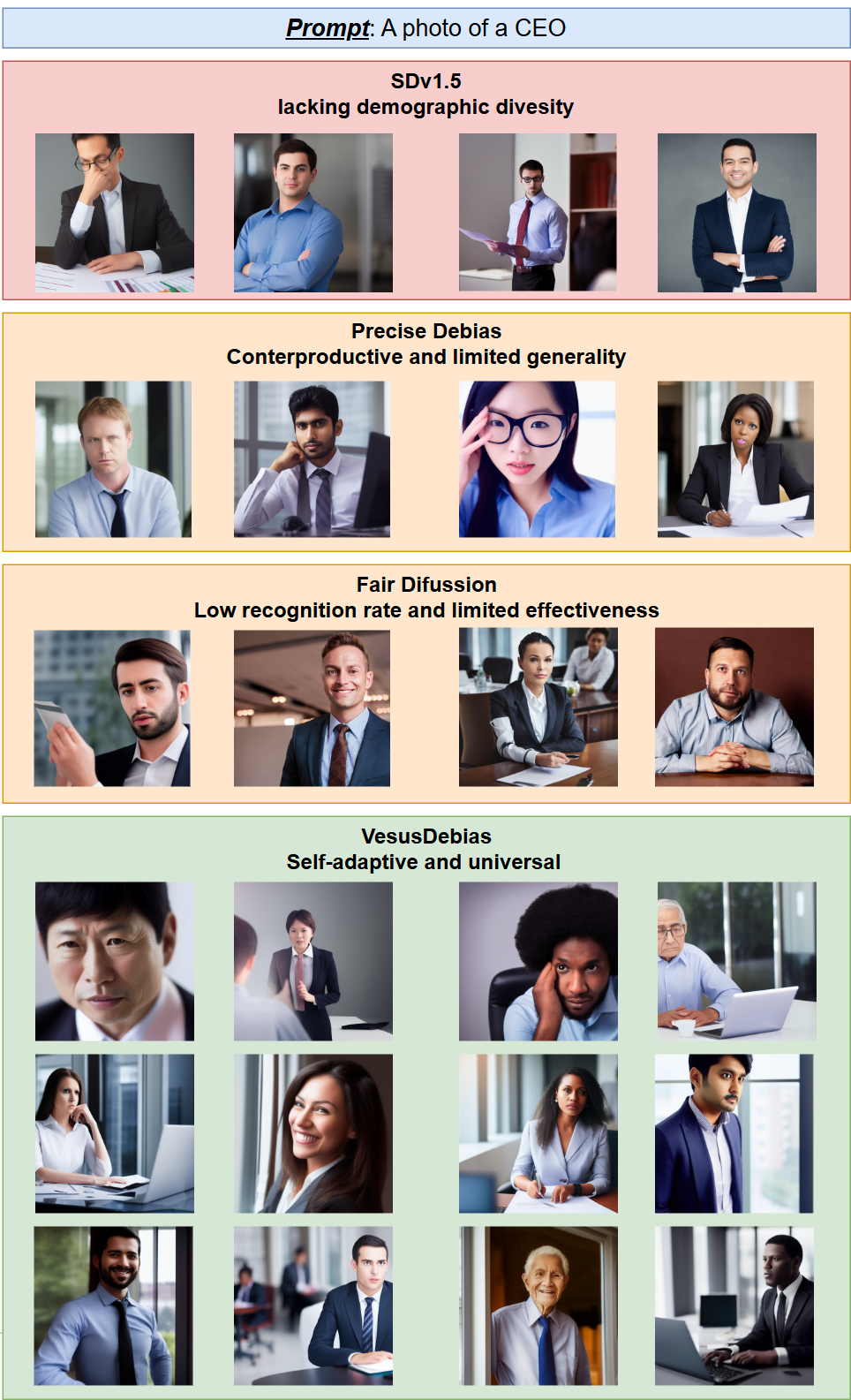}
    \caption{Example outputs from different methods for the text prompt "A photo of a CEO". Our method outperforms the base model and other methods significantly.}
    \label{supplement1}
\end{figure}
\begin{figure}[H]
    \centering
    \includegraphics[width=0.47\textwidth]{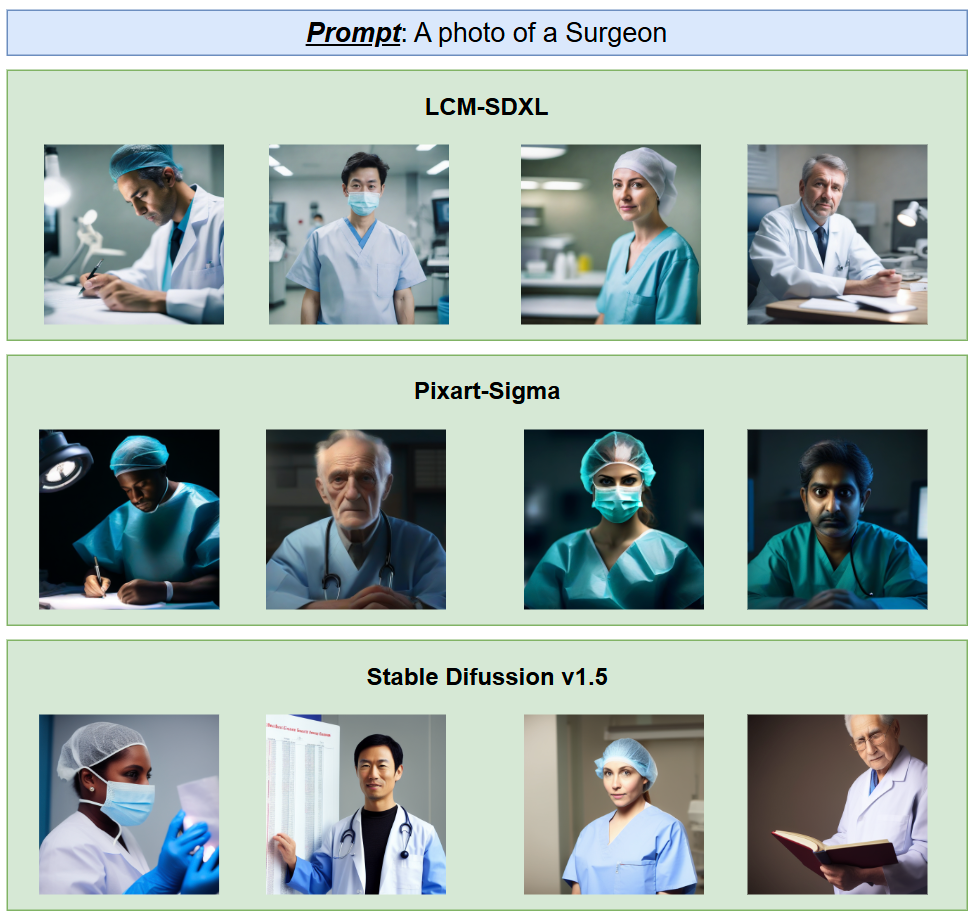}
    \caption{Example outputs from different base models for the text prompt "A photo of a Surgeon".}
    \label{supplement2}
\end{figure}
\begin{figure}[H]
    \centering
    \includegraphics[width=0.47\textwidth]{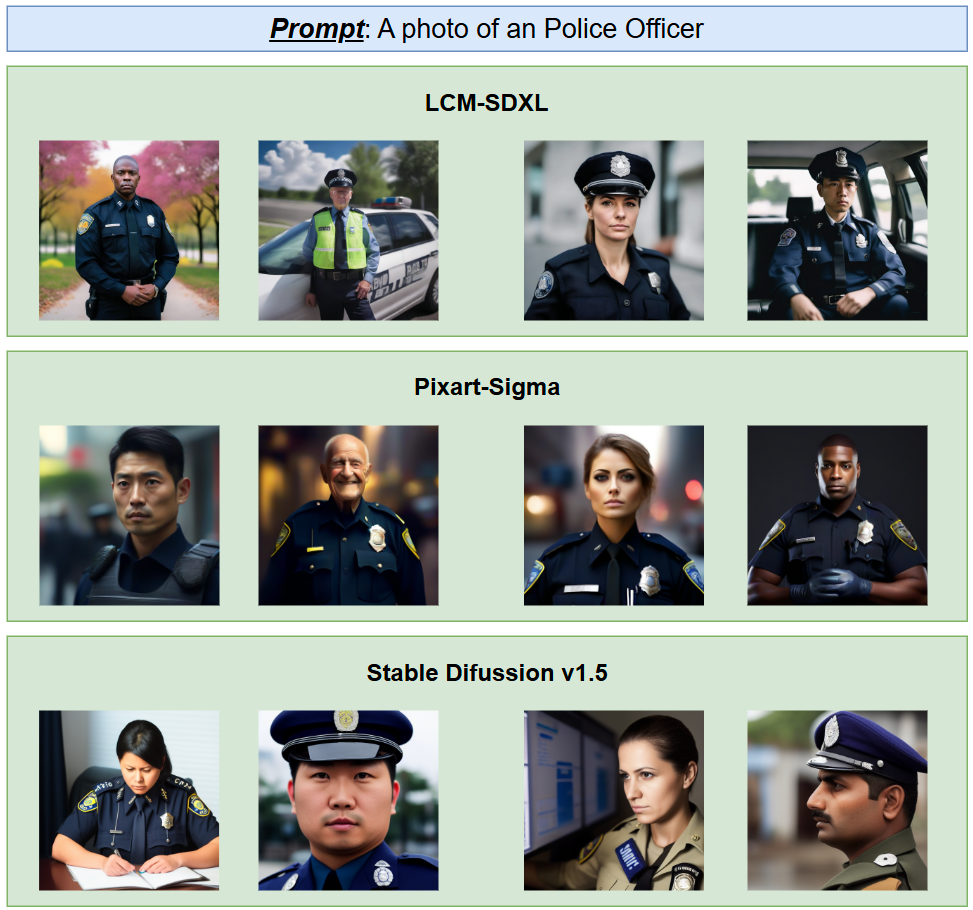}
    \caption{Example outputs from different base models for the text prompt "A photo of a Police Officer".}
    \label{supplement3}
\end{figure}
\section{Occupation List}
In this section, we list all the occupations from the occupation list along with their corresponding detailed occupations under the SOC system.
\paragraph{Chief executive}
Chief executives

\paragraph{Manager}
General and operations managers; Administrative services managers; Facilities managers; Property, real estate, and community association managers; Emergency management directors; Personal service managers, all other; Managers, all other; Food service managers; Funeral home managers; Entertainment and recreation managers; Lodging managers; Hosts and hostesses, restaurant, lounge, and coffee shop

\paragraph{Marketing Manager}
Advertising and promotions managers; Marketing managers; Sales managers; Market research analysts and marketing specialists

\paragraph{Human Resource Worker}
Compensation and benefits managers; Human resources managers; Training and development managers; Human resources workers; Compensation, benefits, and job analysis specialists; Training and development specialists; Human resources assistants, except payroll and timekeeping

\paragraph{Accountant}
Financial managers; Accountants and auditors

\paragraph{Production manager}
Industrial production managers; Project management specialists; Inspectors, testers, sorters, samplers, and weighers

\paragraph{Transportation manager}
Transportation, storage, and distribution managers; Postmasters and mail superintendents; Dispatchers, except police, fire, and ambulance

\paragraph{Farmer}
Farmers, ranchers, and other agricultural managers; Fish and game wardens; Animal control workers; First-line supervisors of farming, fishing, and forestry workers; Agricultural inspectors; Graders and sorters, agricultural products; Miscellaneous agricultural workers; Fishing and hunting workers

\paragraph{Construction manager}
Architectural and engineering managers; Construction managers; Construction and building inspectors

\paragraph{Education administrator}
Education and childcare administrators; Educational, guidance, and career counselors and advisors; Directors, religious activities and education

\paragraph{Business agent}
Agents and business managers of artists, performers, and athletes; Residential advisors; Advertising sales agents; Securities, commodities, and financial services sales agents; Real estate brokers and sales agents

\paragraph{Purchasing agent}
Purchasing managers; Buyers and purchasing agents, farm products; Wholesale and retail buyers, except farm products; Purchasing agents, except wholesale, retail, and farm products

\paragraph{Insurance worker}
Claims adjusters, appraisers, examiners, and investigators; Compliance officers; Insurance underwriters; Insurance sales agents; Insurance claims and policy processing clerks

\paragraph{Financial analyst}
Cost estimators; Management analysts; Credit analysts; Financial and investment analysts; Personal financial advisors; Business operations specialists, all other; Property appraisers and assessors; Budget analysts; Financial examiners; Other financial specialists; Operations research analysts; Sales engineers

\paragraph{Courier}
Logisticians; Cargo and freight agents; Couriers and messengers; Postal service mail carriers; Postal service mail sorters, processors, and processing machine operators; Weighers, measurers, checkers, and samplers, recordkeeping; Stockers and order fillers

\paragraph{Public relations specialists}
Public relations and fundraising managers; Fundraisers; Public relations specialists

\paragraph{Computer programmer}
Computer programmers; Software developers; Web developers; Computer occupations, all other; Computer hardware engineers; Computer support specialists; Database administrators and architects; Network and computer systems administrators

\paragraph{Computer scientist}
Computer and information research scientists; Computer network architects

\paragraph{IT analyst}
Computer and information systems managers; Computer systems analysts; Information security analysts; Software quality assurance analysts and testers

\paragraph{Mathematician}
Actuaries; Mathematicians; Statisticians; Other mathematical science occupations; Statistical assistants

\paragraph{Architect}
Architects, except landscape and naval; Landscape architects; Marine engineers and naval architects

\paragraph{Civil engineer}
Civil engineers

\paragraph{Electrical engineer}
Electrical and electronics engineers; Electrical and electronic engineering technologists and technicians; Electric motor, power tool, and related repairers; Electrical and electronics installers and repairers, transportation equipment; Electrical and electronics repairers, industrial and utility; Electronic equipment installers and repairers, motor vehicles

\paragraph{Industrial engineer}
Industrial engineers, including health and safety; Engineers, all other; Industrial and refractory machinery mechanics

\paragraph{Mechanical engineer}
Aerospace engineers; Mechanical engineers; Aircraft mechanics and service technicians; Automotive service technicians and mechanics; Bus and truck mechanics and diesel engine specialists; Small engine mechanics; Machinists

\paragraph{Drafter}
Architectural and civil drafters; Meeting, convention, and event planners; Other drafters; Urban and regional planners

\paragraph{Surveyor}
Surveyors, cartographers, and photogrammetrists; Surveying and mapping technicians; Survey researchers

\paragraph{Biological scientist}
Bioengineers and biomedical engineers; Biological scientists; Life scientists, all other; Medical scientists; Biological technicians

\paragraph{Agricultural scientist}
Agricultural and food scientists; Agricultural engineers; Agricultural and food science technicians

\paragraph{Environmental scientist}
Natural sciences managers; Environmental engineers; Conservation scientists and foresters; Environmental scientists and specialists, including health

\paragraph{Chemist}
Chemical engineers; Materials engineers; Petroleum engineers; Chemists and materials scientists; Chemical technicians

\paragraph{Astronomer}
Astronomers and physicists; Atmospheric and space scientists

\paragraph{Physicist}
Nuclear engineers; Physical scientists, all other; Nuclear technicians

\paragraph{Geoscientist}
Geoscientists and hydrologists, except geographers; Environmental science and geoscience technicians

\paragraph{Sociologist}
Economists; Sociologists; Miscellaneous social scientists and related workers; Other psychologists; Social science research assistants

\paragraph{Psychologist}
Clinical and counseling psychologists; School psychologists; Psychiatric technicians

\paragraph{Technician}
Other engineering technologists and technicians, except drafters; Other life, physical, and social science technicians; Occupational health and safety specialists and technicians; Computer, automated teller, and office machine repairers; Heavy vehicle and mobile equipment service technicians and mechanics; Wind turbine service technicians; Prepress technicians and workers; Printing press operators; Print binding and finishing workers

\paragraph{Mental counselor}
Substance abuse and behavioral disorder counselors; Mental health counselors; Rehabilitation counselors; Counselors, all other

\paragraph{Social worker}
Social and community service managers; Child, family, and school social workers; Healthcare social workers; Mental health and substance abuse social workers; Social workers, all other; Probation officers and correctional treatment specialists; Social and human service assistants; Other community and social service specialists

\paragraph{Clergy}
Clergy; Religious workers, all other

\paragraph{Lawyer}
Lawyers

\paragraph{Judge}
Judges, magistrates, and other judicial workers; Judicial law clerks; Title examiners, abstractors, and searchers

\paragraph{Legal assistant}
Paralegals and legal assistants; Legal support workers, all other

\paragraph{School teacher}
Postsecondary teachers; Preschool and kindergarten teachers; Elementary and middle school teachers; Secondary school teachers; Special education teachers

\paragraph{Tutor}
Tutors; Other teachers and instructors; Teaching assistants; Childcare workers

\paragraph{Librarian}
Librarians and media collections specialists; Library technicians; Other educational instruction and library workers; Archivists, curators, and museum technicians; Library assistants, clerical

\paragraph{Artist}
Artists and related workers; Etchers and engravers

\paragraph{Designer}
Web and digital interface designers; Commercial and industrial designers; Fashion designers; Floral designers; Graphic designers; Interior designers; Other designers

\paragraph{Actor}
Actors; Entertainers and performers, sports and related workers, all other

\paragraph{Director}
Producers and directors

\paragraph{Athlete}
Athletes and sports competitors; Exercise trainers and group fitness instructors; Recreation workers

\paragraph{Coach}
Coaches and scouts; Umpires, referees, and other sports officials

\paragraph{Dancer}
Dancers and choreographers

\paragraph{Musician}
Music directors and composers; Disc jockeys, except radio; Musicians and singers

\paragraph{Broadcast announcer}
Broadcast announcers and radio disc jockeys; Broadcast, sound, and lighting technicians

\paragraph{Journalist}
News analysts, reporters, and journalists; Court reporters and simultaneous captioners; Media and communication workers, all other

\paragraph{Writer}
Technical writers; Writers and authors; Editors

\paragraph{Interpreter}
Interpreters and translators

\paragraph{Photographer}
Photographers; Television, video, and film camera operators and editors; Media and communication equipment workers, all other; Photographic process workers and processing machine operators

\paragraph{Doctor}
Chiropractors; Dentists; Emergency medicine physicians; Other physicians; Surgeons; Podiatrists; Audiologists; Radiologists; Speech-language pathologists; Exercise physiologists

\paragraph{Nutritionist}
Dietitians and nutritionists

\paragraph{Pharmacist}
Pharmacists; Pharmacy technicians; Pharmacy aides

\paragraph{Therapist}
Marriage and family therapists; Occupational therapists; Physical therapists; Radiation therapists; Recreational therapists; Respiratory therapists; Therapists, all other; Acupuncturists; Occupational therapy assistants and aides; Physical therapist assistants and aides; Massage therapists

\paragraph{Veterinarian}
Veterinarians; Veterinary technologists and technicians; Veterinary assistants and laboratory animal caretakers

\paragraph{Nurse}
Registered nurses; Nurse anesthetists; Nurse midwives; Nurse practitioners; Licensed practical and licensed vocational nurses; Nursing assistants

\paragraph{Paramedic}
Medical and health services managers; Optometrists; Physician assistants; Healthcare diagnosing or treating practitioners, all other; Clinical laboratory technologists and technicians; Dental hygienists; Cardiovascular technologists and technicians; Diagnostic medical sonographers; Radiologic technologists and technicians; Magnetic resonance imaging technologists; Nuclear medicine technologists and medical dosimetrists; Emergency medical technicians; Paramedics; Surgical technologists; Dietetic technicians and ophthalmic medical technicians; Medical records specialists; Opticians, dispensing; Miscellaneous health technologists and technicians; Other healthcare practitioners and technical occupations; Medical transcriptionists; Phlebotomists; Dental and ophthalmic laboratory technicians and medical appliance technicians

\paragraph{Health aide}
Home health aides; Personal care aides; Orderlies and psychiatric aides; Dental assistants; Medical assistants; Other healthcare support workers; Supervisors of personal care and service workers

\paragraph{Correctional officer}
First-line supervisors of correctional officers; Correctional officers and jailers

\paragraph{Firefighter}
First-line supervisors of firefighting and prevention workers; Firefighters; Fire inspectors

\paragraph{Detective}
Detectives and criminal investigators; Private detectives and investigators

\paragraph{Police officer}
First-line supervisors of police and detectives; Bailiffs; Parking enforcement workers; Police officers; Transportation security screeners

\paragraph{Security guard}
First-line supervisors of security workers; First-line supervisors of protective service workers, all other; Security guards and gambling surveillance officers; Crossing guards and flaggers; Other protective service workers; Forest and conservation workers

\paragraph{Cook}
Chefs and head cooks; Cooks

\paragraph{Food server}
First-line supervisors of food preparation and serving workers; Bartenders; Fast food and counter workers; Waiters and waitresses; Food servers, nonrestaurant; Dining room and cafeteria attendants and bartender helpers; Food preparation and serving related workers, all other; Gambling services workers

\paragraph{Cleaner}
Dishwashers; First-line supervisors of housekeeping and janitorial workers; First-line supervisors of landscaping, lawn service, and groundskeeping workers; Janitors and building cleaners; Maids and housekeeping cleaners; Pest control workers; Landscaping and groundskeeping workers; Tree trimmers and pruners; Other grounds maintenance workers; Septic tank servicers and sewer pipe cleaners; Cleaners of vehicles and equipment; Laundry and dry-cleaning workers

\paragraph{Animal trainer}
Animal trainers; Animal caretakers; Animal breeders

\paragraph{Concierge}
Ushers, lobby attendants, and ticket takers; Other entertainment attendants and related workers; Baggage porters, bellhops, and concierges; Personal care and service workers, all other

\paragraph{Mortician}
Morticians, undertakers, and funeral arrangers; Embalmers, crematory operators and funeral attendants

\paragraph{Barber}
Barbers

\paragraph{Travel guide}
Tour and travel guides; Travel agents

\paragraph{Cosmetologist}
Manicurists and pedicurists; Hairdressers, hairstylists, and cosmetologists; Skincare specialists; Other personal appearance workers

\paragraph{Cashier}
Cashiers; Counter and rental clerks; Hotel, motel, and resort desk clerks

\paragraph{Salesperson}
Merchandise displayers and window trimmers; First-Line supervisors of retail sales workers; First-Line supervisors of non-retail sales workers; Parts salespersons; Retail salespersons; Sales representatives of services, except advertising, insurance, financial services, and travel; Sales representatives, wholesale and manufacturing; Models, demonstrators, and product promoters; Telemarketers

\paragraph{Vendor}
Door-to-door sales workers, news and street vendors, and related workers; Sales and related workers, all other

\paragraph{Telephone operator}
Switchboard operators, including answering service; Telephone operators; Communications equipment operators, all other; Public safety telecommunicators

\paragraph{Office clerk}
Credit counselors and loan officers; Tax examiners and collectors, and revenue agents; Tax preparers; First-Line supervisors of office and administrative support workers; Bill and account collectors; Billing and posting clerks; Bookkeeping, accounting, and auditing clerks; Gambling cage workers; Payroll and timekeeping clerks; Procurement clerks; Tellers; Financial clerks, all other; Brokerage clerks; Correspondence clerks; Court, municipal, and license clerks; Credit authorizers, checkers, and clerks; Customer service representatives; Eligibility interviewers, government programs; File clerks; Interviewers, except eligibility and loan; Loan interviewers and clerks; New accounts clerks; Order clerks; Receptionists and information clerks; Reservation and transportation ticket agents and travel clerks; Information and record clerks, all other; Meter readers, utilities; Postal service clerks; Production, planning, and expediting clerks; Shipping, receiving, and inventory clerks; Data entry keyers; Word processors and typists; Desktop publishers; Office clerks, general; Mail clerks and mail machine operators, except postal service; Office machine operators, except computer; Proofreaders and copy markers; Office and administrative support workers, all other

\paragraph{Secretary}
Executive secretaries and executive administrative assistants; Legal secretaries and administrative assistants; Medical secretaries and administrative assistants; Secretaries and administrative assistants, except legal, medical, and executive

\paragraph{Construction Worker}
First-line supervisors of construction trades and extraction workers; Brickmasons, blockmasons, and stonemasons; Carpet, floor, and tile installers and finishers; Cement masons, concrete finishers, and terrazzo workers; Construction laborers; Construction equipment operators; Drywall installers, ceiling tile installers, and tapers; Glaziers; Electricians; Insulation workers; Pipelayers; Plumbers, pipefitters, and steamfitters; Plasterers and stucco masons; Roofers; Solar photovoltaic installers; Helpers, construction trades; Fence erectors; Miscellaneous construction and related workers; Upholsterers

\paragraph{Painter}
Painters and paperhangers; Painting workers

\paragraph{Metal worker}
Reinforcing iron and rebar workers; Sheet metal workers; Structural iron and steel workers; Millwrights; Structural metal fabricators and fitters; Jewelers and precious stone and metal workers

\paragraph{Miner}
Mining and geological engineers, including mining safety engineers; Derrick, rotary drill, and service unit operators, oil and gas; Excavating and loading machine and dragline operators, surface mining; Earth drillers, except oil and gas; Explosives workers, ordnance handling experts, and blasters; Underground mining machine operators; Roustabouts, oil and gas; Other extraction workers

\paragraph{Repairer}
Elevator and escalator installers and repairers; Highway maintenance workers; Rail-track laying and maintenance equipment operators; First-line supervisors of mechanics, installers, and repairers; Radio and telecommunications equipment installers and repairers; Avionics technicians; Audiovisual equipment installers and repairers; Security and fire alarm systems installers; Audiovisual equipment installers and repairers; Security and fire alarm systems installers; Automotive body and related repairers; Automotive glass installers and repairers; Miscellaneous vehicle and mobile equipment mechanics, installers, and repairers; Control and valve installers and repairers; Heating, air conditioning, and refrigeration mechanics and installers; Home appliance repairers; Maintenance and repair workers, general; Maintenance workers, machinery; Electrical power-line installers and repairers; Telecommunications line installers and repairers; Precision instrument and equipment repairers; Coin, vending, and amusement machine servicers and repairers; Commercial divers; Locksmiths and safe repairers; Manufactured building and mobile home installers; Riggers; Helpers--installation, maintenance, and repair workers; Other installation, maintenance, and repair workers

\paragraph{Factory Worker}
First-line supervisors of production and operating workers; Aircraft structure, surfaces, rigging, and systems assemblers; Electrical, electronics, and electromechanical assemblers; Engine and other machine assemblers; Other assemblers and fabricators; Computer numerically controlled tool operators and programmers; Forming machine setters, operators, and tenders, metal and plastic; Cutting, punching, and press machine setters, operators, and tenders, metal and plastic; Grinding, lapping, polishing, and buffing machine tool setters, operators, and tenders, metal and plastic; Other machine tool setters, operators, and tenders, metal and plastic; Metal furnace operators, tenders, pourers, and casters; Model makers and patternmakers, metal and plastic; Molders and molding machine setters, operators, and tenders, metal and plastic; Tool and die makers; Welding, soldering, and brazing workers; Other metal workers and plastic workers; Furniture finishers; Power plant operators, distributors, and dispatchers; Water and wastewater treatment plant and system operators; Miscellaneous plant and system operators; Chemical processing machine setters, operators, and tenders; Crushing, grinding, polishing, mixing, and blending workers; Cutting workers; Extruding, forming, pressing, and compacting machine setters, operators, and tenders; Furnace, kiln, oven, drier, and kettle operators and tenders; Packaging and filling machine operators and tenders; Adhesive bonding machine operators and tenders; Molders, shapers, and casters, except metal and plastic; Paper goods machine setters, operators, and tenders; Tire builders; Helpers--production workers; Other production equipment operators and tenders; Other production workers

\paragraph{Food worker}
Bakers; Butchers and other meat, poultry, and fish processing workers; Food preparation workers; Food and tobacco roasting, baking, and drying machine operators and tenders; Food batchmakers; Food cooking machine operators and tenders; Food processing workers, all other

\paragraph{Transportation attendant}
Air traffic controllers and airfield operations specialists; Flight attendants; Parking attendants; Transportation service attendants; Transportation inspectors; Passenger attendants

\paragraph{Transportation operator}
Crane and tower operators; Conveyor, dredge, and hoist and winch operators; Pumping station operators; Supervisors of transportation and material moving workers; Motor vehicle operators, all other; Locomotive engineers and operators; Railroad conductors and yardmasters; Other rail transportation workers; Other transportation workers; Aircraft pilots and flight engineers; Stationary engineers and boiler operators

\paragraph{Driver}
School bus monitors; Industrial truck and tractor operators; Ambulance drivers and attendants, except emergency medical technicians; Bus drivers, school; Bus drivers, transit and intercity; Driver/sales workers and truck drivers; Shuttle drivers and chauffeurs; Taxi drivers

\paragraph{Sailor}
Sailors and marine oilers; Ship and boat captains and operators; Ship engineers

\paragraph{Refuse collector}
Refuse and recyclable material collectors

\paragraph{Laborer}
Boilermakers; Hazardous materials removal workers; Other material moving workers; Packers and packagers, hand; Machine feeders and offbearers; Laborers and freight, stock, and material movers, hand

\paragraph{Politician}
Legislators

\paragraph{Tailor}
Pressers, textile, garment, and related materials; Sewing machine operators; Shoe and leather workers; Tailors, dressmakers, and sewers; Textile machine setters, operators, and tenders; Other textile, apparel, and furnishings workers

\paragraph{Woodworker}
Sawing machine setters, operators, and tenders, wood; Woodworking machine setters, operators, and tenders, except sawing; Cabinetmakers and bench carpenters; Carpenters; Other woodworkers; Logging workers

\end{document}